\documentclass[letterpaper, 10 pt, conference]{ieeeconf}  

\IEEEoverridecommandlockouts                              

\usepackage[english]{babel} 
\usepackage{graphicx,adjustbox}
\usepackage{cite}
\usepackage{amsmath,amsfonts,amssymb,amsbsy}
\usepackage{pifont}
\usepackage{algorithm}
\usepackage{algorithmic}
\usepackage{comment}
\usepackage{caption}
\usepackage{subcaption}
\usepackage{float}    
\usepackage{placeins} 
\usepackage{svg}
\usepackage[T1]{fontenc}
\usepackage{flushend} 
\usepackage{booktabs,threeparttable,multirow,array}

\usepackage[raiselinks=true,
            bookmarks=true,
            bookmarksopenlevel=1,
            bookmarksopen=true,
            bookmarksnumbered=true,
            hyperindex=true,
            plainpages=false,
            pdfpagelabels=true,
            pdfpagelabels=true,
            pdfborder={0 0 0.5}]{hyperref}
\usepackage[capitalise]{cleveref}

\definecolor{custom-link-color}{rgb}{0.2 0.45 1.0} 
\definecolor{custom-cite-color}{rgb}{0.1 0.45 0.05} 
\hypersetup{                
    pdfauthor={"Marius Baden and Ahmed Abouelazm"},
    pdftitle={"Prior Knowledge Integration for Feasible and Interpretable Trajectory Prediction"},
    colorlinks=true,
    linkcolor = cyan,	
    citecolor = green,
}
\crefname{figure}{Fig.}{Figs.}

\usepackage{eso-pic}
\newcommand{\mycopyrighttext}{%
  \footnotesize
  \noindent
  \textcopyright~2025 IEEE. Personal use of this material is permitted. Permission from IEEE must be obtained for all other uses, in any current or future media, including reprinting/republishing this material for advertising or promotional purposes, creating new collective works, for resale or redistribution to servers or lists, or reuse of any copyrighted component of this work in other works.\\
  IEEE 36th Intelligent Vehicles Symposium (IV 2025) - 22-25 June, 2025.
}

\AtBeginDocument{%
  \AddToShipoutPicture*{%
    \AtPageUpperLeft{%
      \put(55, -40){
        \parbox{\dimexpr\textwidth-4pt\relax}{\raggedright \mycopyrighttext}
      }%
    }
  }
}

\title{\LARGE \bf 
TPK: Trustworthy Trajectory Prediction Integrating Prior Knowledge For Interpretability and Kinematic Feasibility
}

\author{Marius Baden\textsuperscript{\textasteriskcentered}$^{1,2}$, Ahmed Abouelazm\textsuperscript{\textasteriskcentered}$^{1}$, Christian Hubschneider$^{1,2}$, \\ Yin Wu$^{2,3}$, Daniel Slieter$^{3}$, and J. Marius Zöllner$^{1,2}$
\thanks{\textasteriskcentered~These authors contributed equally to this work}%
\thanks{$^{1}$Authors are with the FZI Research Center for Information Technology, Germany
        {\tt\small abouelazm@fzi.de}}%
\thanks{$^{2}$Authors are with the Karlsruhe Institute of Technology, Germany}%
\thanks{$^{3}$Authors are with CARIAD SE, Germany}%
}

\begin{document}
\maketitle
\thispagestyle{empty}
\pagestyle{empty}

\begin{abstract}
    Trajectory prediction is crucial for autonomous driving, enabling vehicles to navigate safely by anticipating the movements of surrounding road users. However, current deep learning models often lack trustworthiness as their predictions can be physically infeasible and illogical to humans. 
    To make predictions more trustworthy, recent research has incorporated prior knowledge, like the social force model for modeling interactions and kinematic models for physical realism. However, these approaches focus on priors that suit either vehicles or pedestrians and do not generalize to traffic with mixed agent classes.
    We propose incorporating interaction and kinematic priors of all agent classes--vehicles, pedestrians, and cyclists with class-specific interaction layers to capture agent behavioral differences.
    To improve the interpretability of the agent interactions, we introduce DG-SFM, a rule-based interaction importance score that guides the interaction layer. 
    To ensure physically feasible predictions, we proposed suitable kinematic models for all agent classes with a novel pedestrian kinematic model.
    We benchmark our approach on the Argoverse 2 dataset, using the state-of-the-art transformer HPTR as our baseline.
    Experiments demonstrate that our method improves interaction interpretability, revealing a correlation between incorrect predictions and divergence from our interaction prior.
    Even though incorporating the kinematic models causes a slight decrease in accuracy, they eliminate infeasible trajectories found in the dataset and the baseline model.
    Thus, our approach fosters trust in trajectory prediction as its interaction reasoning is interpretable, and its predictions adhere to physics. 
    

    \begin{keywords}
        Trajectory Prediction, Motion Planning
    \end{keywords} 
\end{abstract}
\section{Introduction}
\label{sec:Introduction}
 
Road injuries are the leading cause of death among children and young adults worldwide, according to the World Health Organization \cite{who2023}, highlighting the urgent need for safer roads. Autonomous driving presents a promising solution, with trajectory prediction playing a critical role by enabling vehicles to anticipate the movements of surrounding agents and safely navigate roads \cite{ding2023}. 

\begin{figure}[t]
    \centering
    \begin{subfigure}[b]{0.98\linewidth}
    \includegraphics[trim={0cm 0.60cm 0cm 0.50cm},clip,width=\linewidth]{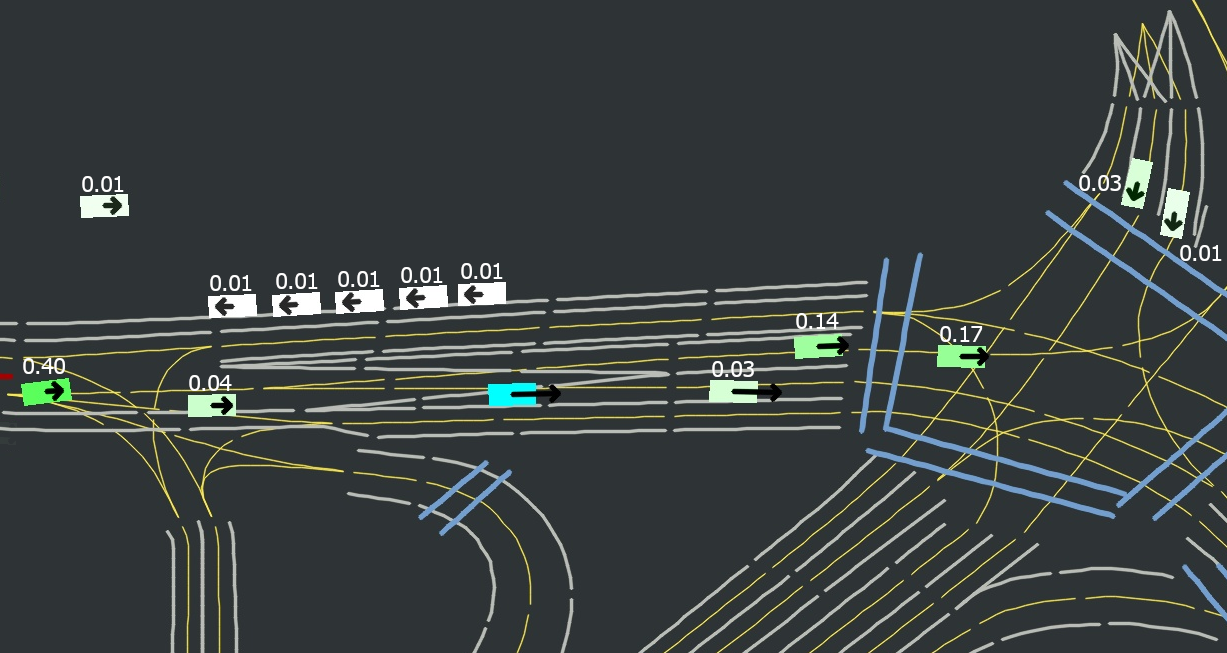}
    \caption{
        HPTR's \cite{hptr} agent attention \textbf{contradicts} human reasoning, as the agent in front of the focal agent barely influences the prediction, while an agent far behind has a significant impact. 
    }
    \label{fig:face_figure}
    \end{subfigure}%
    \vspace{5pt}
    \begin{subfigure}[b]{0.98\linewidth}
    \includegraphics[trim={0cm 0.60cm 0cm 0.5cm},clip,width=\linewidth]{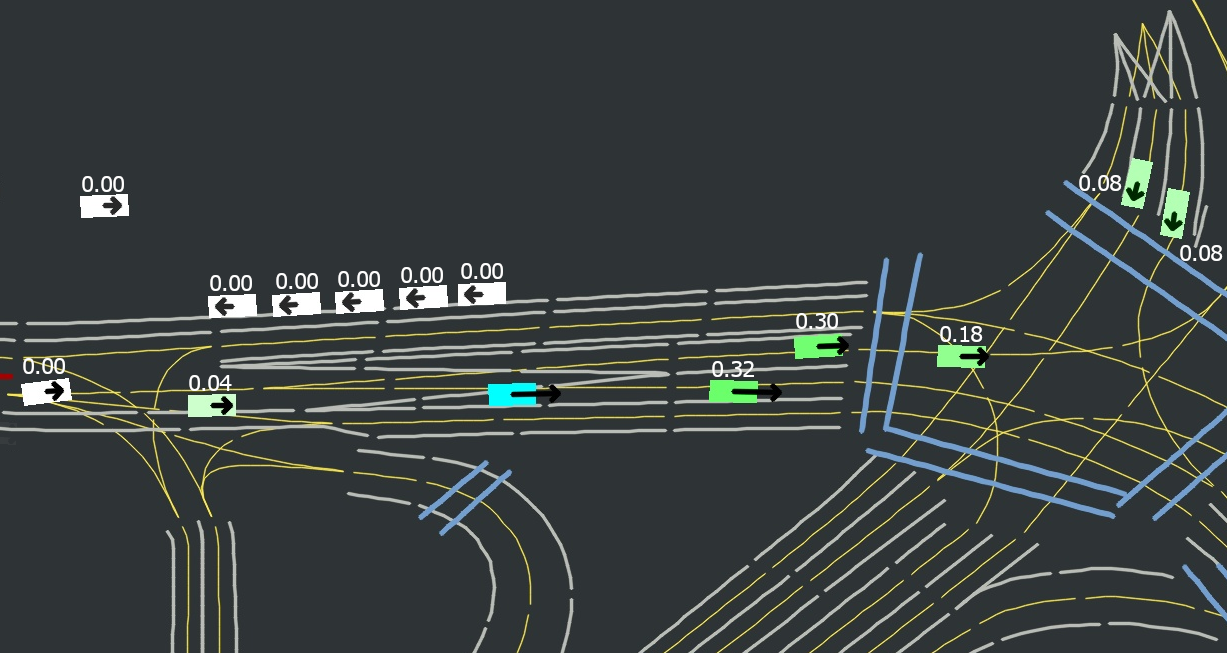}
    \caption{
        Our proposed model \textbf{corrects} HPTR's untrustworthy interaction reasoning by assigning attention scores that align with human intuition, guided by our novel DG-SFM prior.
    }
    \label{fig:dg-sfm_attn}
    \end{subfigure}
    \caption{Comparison of learned agent attention in HPTR and our proposed model on a scene from the ArgoVerse 2 dataset. The focal agent is highlighted in cyan, with numbers indicating the attention score per agent and black arrows representing agent velocities.}
\end{figure}

While neural networks have significantly advanced trajectory prediction by learning complex patterns from large datasets  \cite{ding2023}, their data-driven nature can lead to illogical or flawed predictions that deviate from human reasoning. For example, as illustrated in \cref{fig:face_figure}, state-of-the-art models may predict trajectories influenced by high importance assigned to agents that are irrelevant to the scenario.
These limitations undermine the trustworthiness of trajectory prediction systems, especially in rare but safety-critical scenarios, ultimately hindering the practical application of these methods in autonomous driving \cite{carrasco2024}.

To tackle these trustworthiness challenges, recent research has explored hybrid prediction methods by integrating expert knowledge into neural networks, referred to as a \textit{prior}. Models such as the social force model (SFM) have been employed to capture agent interactions \cite{sfm95}, while kinematic models are utilized to ensure physical feasibility \cite{dkm}. 
However, existing approaches of interaction priors are limited to pedestrian interactions and often integrate the prior into the network without providing a clear justification for the chosen integration methods. Similarly, research on physically feasible prediction has focused on vehicles and cyclists, while pedestrians are left mostly unconstrained.

Rather than solely pursuing further improvements in the already high accuracy of state-of-the-art trajectory prediction models, we follow the argumentation of~\cite{carrasco2024} that enhancing their trustworthiness is the critical next step. To achieve this, we propose a novel interaction prior applicable to all agent classes, compare two integration methods, and guarantee kinematic feasibility specific to each agent type. Integrating the interaction prior into agent-to-agent attention aligns the model prediction with human intuition, enhancing its interpretability, as shown in \Cref{fig:dg-sfm_attn}. Additionally, enforcing kinematic feasibility through agent-specific kinematic layers ensures predictions adhere to physical laws.
Overall, we introduce three key contributions to enhance the trustworthiness of trajectory prediction networks:
\begin{itemize}
    \item \textbf{A novel interaction prior} for agents in traffic scenarios with heterogeneous classes.
    \item \textbf{Enhanced interpretability of agent interactions} by integrating the prior into the attention mechanism and comparing different integration methods.
    \item \textbf{Guaranteed kinematic feasibility for all agent classes} through kinematic layers injected directly into the network, including a novel kinematic layer for pedestrians.
\end{itemize}

\section{Related Work}
\label{sec:sota}
Hybrid prediction models combine the flexibility of deep learning with the interpretability of knowledge-based methods to address feasibility and generalization challenges \cite{ding2023}. These models often emphasize quality metrics beyond accuracy, such as physical feasibility \cite{golchoubian2023, ptnet}. This section examines the integration of prior knowledge into trajectory prediction models, focusing on two key areas: \textit{interaction priors} for interpretability and \textit{kinematic priors} for feasibility.

\subsection{Interaction Priors}
\label{sec:sota:interaction_prior_integration}
Graph- and transformer-based models typically model social interactions through an attention mechanism \cite{wang2022} to assign \textit{interaction importance scores}, which quantify the influence of neighboring agents on the focal agent’s predictions~\cite{xhgp}. Incorporating prior knowledge into attention mechanisms is an emerging research topic, leveraging their inherent interpretability and critical role in shaping the embeddings that drive the model’s predictions.

SFM \cite{sfm95}, widely used for rule-based pedestrian trajectory prediction, represents social interactions as repulsive and attractive forces between agents. While integrated into some deep learning models, it is typically used as an additional input rather than as an interaction prior \cite{forceformer}.
These models often focus on accuracy improvements without analyzing how the SFM inputs affect the predictions. Approaches proposed in \cite{sfm-nn} embed the SFM into neural network architecture, treating its equations as trainable layers. Numerous extensions of the SFM aim to enhance the SFM for specific applications like simulating traffic in shared open spaces~\cite{farina2017walking}. However, adopting the SFM's social forces as interaction importance scores introduces notable challenges. For instance, SFM does not consider the focal agent's heading or speed, resulting in symmetric force scores for agents positioned both in front and behind.

Another social interaction model, SKGACN \cite{skgacn} generates interaction scores based on spatial and motion-related factors among pedestrians. These scores are calculated using the Euclidean distance, relative velocity, and heading, then normalized with a softmax function. SKGACN integrates these scores into its attention mechanism by multiplying them with learned agent embeddings \cite{skgacn}. Compared to the SFM, SKGACN is simpler and accounts for velocity and heading. However, its design is specific to pedestrians, limiting its use for vehicle or mixed traffic trajectory prediction.

Once an interaction prior is established, integrating it into attention mechanisms poses a key challenge. While prior integration is relatively new to trajectory prediction, it has been explored in other fields of deep learning. We identified four prevalent methods in the literature. 
The \textbf{contextual prior} methods treat the prior as an input by concatenating it with the network's embedding \cite{margatina2019, srgat}. 
While highly flexible, this approach might lead the model to ignore the prior training. 
Another approach involves \textbf{regularization loss}, adding a term to align attention scores with the prior \cite{bam}.
This ensures the prior's influence but increases training complexity.
\textbf{Multiply-and-renormalize} methods incorporate the prior by multiplying it with learned attention scores and renormalizing them \cite{bert-sim, skgacn}. 
This makes the prior’s impact explicit but can mask out the network contribution when the prior is close to zero. 
Finally, the \textbf{gating} methods use a learnable gate to balance the prior and learned attention \cite{ame}. 
While more explicit than contextual priors and more flexible than multiply-and-renormalize, it shares the drawback of the contextual approach, where the prior may be ignored unless reinforced by an additional loss function. 

\subsection{Kinematic Priors}
\label{sec:sota:feasibility}
Recent works have tackled the challenge of ensuring kinematic feasibility in trajectory prediction. SafetyNet \cite{safetynet} uses post-processing to detect infeasible trajectories and applies a rule-based fallback to ensure kinematic feasibility. However, this fallback reduces accuracy, as it relies on rule-based predictions, which lack the precision of data-driven approaches. 
In contrast, action-space prediction focuses on predicting control inputs such as acceleration and steering angles. These approaches~\cite{dkm, trajectron++} employ deterministic and differentiable kinematic models to transform control inputs into predicted trajectories. Action-space prediction is modular, enforcing kinematic constraints by adding a kinematic model without altering the core network architecture~\cite{dkm}. Additionally, it accommodates different kinematic models tailored to various agent types~\cite{trajectron++}. 
The two-axle vehicle model and the bicycle model are widely applicable for vehicles and cyclists. The two-axle vehicle model provides the most accurate representation of vehicle dynamics~\cite{dkm} and is particularly suited for scenarios requiring precise physical modeling. However, this accuracy involves the estimation of numerous agent-specific parameters. These parameters, often estimated in real-time from tracking data, can introduce errors and reduce the model's precision. A common simplification is the bicycle model~\cite{ssp-asp}. 
This model balances accuracy and simplicity, offering an effective approximation.

For pedestrian trajectory prediction, the choice of a kinematic model is less standardized. Some approaches forgo kinematic models entirely~\cite{dkm}. This allows maximum flexibility to capture diverse behaviors but lacks feasibility guarantees. The single integrator model uses 2D velocity as control inputs while imposing a maximum speed limit~\cite{trajectron++}. This approach introduces continuity in movement and captures local changes effectively, but still permits physically infeasible accelerations. The unicycle model is adapted for pedestrian prediction in~\cite{rudenko2020}, 
as some studies suggest pedestrians in traffic exhibit non-holonomic motion \cite{camara2021}. 
This model enforces smooth turns and avoids abrupt directional changes. Its non-holonomic constraints on orientation changes are unnatural for pedestrians and may limit its ability to capture their movement accurately.

\begin{figure*}[t]
    \centering
    \includegraphics[trim={0.47cm 6.4cm 0.2cm 1.25cm},clip,width=\textwidth]{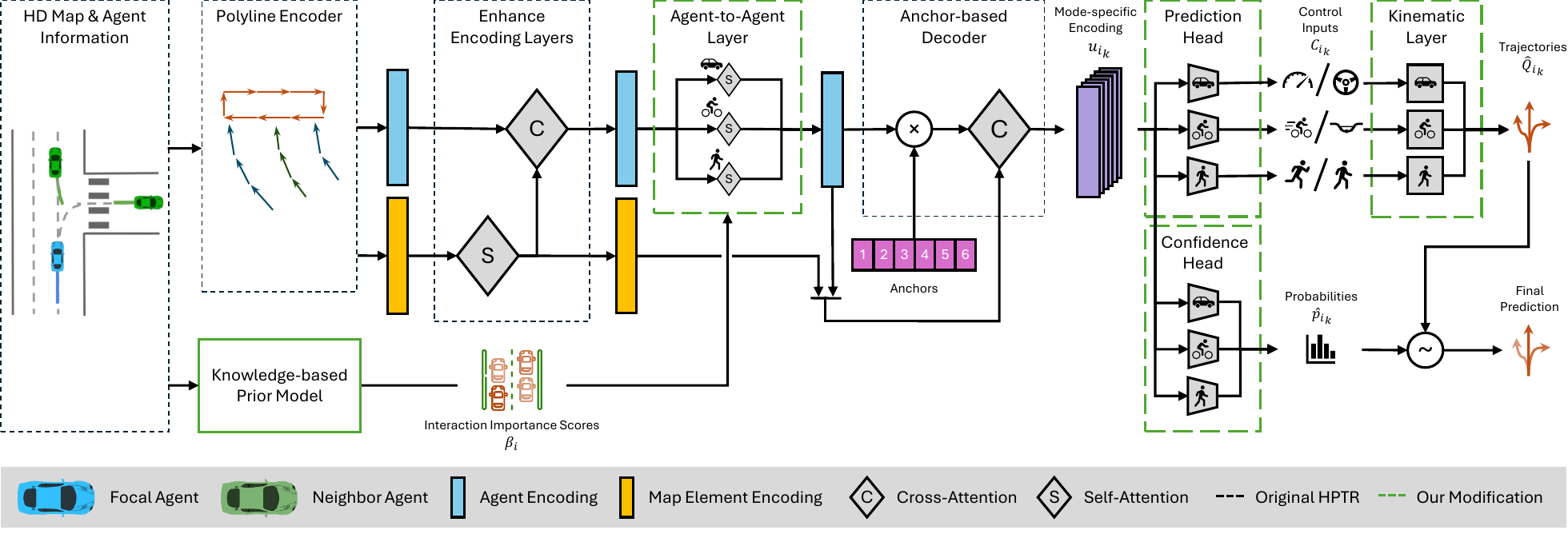}
    \caption{
        The scene is first encoded in agent and map element embeddings and enhanced by transformer layers. 
        Then, the prior-integrated and class-specific agent-to-agent layer captures interactions between agents.
        Next, transformer layers generate mode-specific embeddings based on the agent embeddings.
        Subsequently, class-specific heads predict a sequence of control inputs and a confidence value per mode.
        Finally, a kinematic layer deterministically calculates the predicted trajectories resulting from the control inputs.
    }
    \vspace{-0.5cm}
    \label{fig:system_overview}
\end{figure*}

\section{Methodology}
This section details our approach to improving the trustworthiness and interpretability of trajectory prediction networks, as illustrated in \Cref{fig:system_overview}. We leverage HPTR \cite{hptr} as our backbone, a state-of-the-art transformer-based network known for its efficiency and extensibility. To address the limitations highlighted in ~\Cref{sec:sota}, we propose enhancing encoder interpretability by integrating a novel interaction prior applicable to all agent classes and improving decoder feasibility through a kinematic model, with special emphasis on a novel pedestrian-specific kinematic model.

\subsection{Interpretable Encoder}

The initial layers of our model follow HPTR \cite{hptr}. Scene agents and map information are represented as polylines and encoded using a polyline encoder, as proposed in VectorNet \cite{vectornet}. These polyline embeddings are subsequently refined through multiple transformer layers. To improve the network's interpretability and trustworthiness, we introduce a class-specific agent-to-agent interaction layer guided by a novel prior, which is incorporated before HPTR's anchor-based decoder. This design choice places agent-to-agent attention as the final stage of the encoder, ensuring it has the most prominent influence on the agent embedding passed to the decoder, which is sensible since interactions are a key challenge in trajectory prediction \cite{zhang2022explainable, golchoubian2023}, especially in dense urban environments. Furthermore, this design enforces trustworthy attention in the final encoder stage while keeping earlier stages unconstrained.
\subsubsection{Class-Specific Agent-to-Agent Layer}
\label{sec:method:interpretable:ci}
Interactions between different agent classes are inherently complex \cite{wang2022}.
Hence, similar to \cite{xhgp}, we argue that it is challenging for a network to effectively learn the distinct behaviors of each agent class solely from an agent class label. 
To address this, we introduce a separate instance of the agent-to-agent layer for each agent class. Thus, each instance can act as an expert tailored to the unique interaction patterns of each class. 
Furthermore, class-specific instances mitigate the effects of class imbalance in the datasets \cite{argoverse2, womd}. Without them, a single interaction layer would focus primarily on vehicle interactions, neglecting the underrepresented cyclists and pedestrians classes. We implement each instance using a KNARPE layer \cite{hptr}, an efficient transformer layer that computes attention between an agent's embedding and the embeddings of its $K$-nearest neighboring agents.

\subsubsection{Interaction Prior}
\label{sec:method:enc:interaction_priors}
As previously mentioned, the proposed agent-to-agent layer is guided by an interaction prior, which uses a rule-based model to calculate an \textit{interaction importance score} for surrounding agents.
    SKGACN \cite{skgacn} is a natural choice of prior since it is the only prior addressing interaction importance in trajectory prediction.
    However, SKGACN is specifically designed for pedestrian interactions, making its effectiveness in mixed traffic unclear.
    To overcome this limitation, we introduce a novel prior, \textit{directed-gradient social force model} (DG-SFM), which aligns more closely with human intuition. 
    The intuition behind DG-SFM is that the importance of a neighbor's interaction is determined by its speed and the rate at which it approaches the focal agent. For example, a vehicle rapidly approaching from behind should be considered highly important, as it poses a potential risk of collision. 
    In contrast, a parked vehicle on the side of the road behind the focal agent should have little to no importance, as an interaction is highly unlikely after passing it.

   
    To formalize this intuition, we build upon the SFM by incorporating its notion of elliptical social repulsive potentials \cite{sfm95} to define the personal space of a focal agent \(i\).
    In this formulation, a higher potential value assigned to an agent \(j\) within focal agent \(i\)'s personal space signifies a greater level of risk, reflecting the importance of their interaction. An agent \(j\) can be significant for interactions for two reasons.
    
    Firstly, when agent \(j\) enters the focal agent \(i\)'s personal space, it becomes crucial for the focal agent to monitor \(j\)'s actions in order to anticipate potential interactions. 
    \cref{eq:beta_a} defines the interaction importance \(\beta_{ij}^{\, A}\) assigned from the focal agent \(i\) to agent \(j\). Here, \(r_i = [x_i, y_i]^T\) and \(v_i = [v_{x_i}, v_{y_i}]^T\) represent the 2D position and velocity of agent \(i\) at the last observation time \(T_o\), respectively.
    \begin{equation}
        \beta_{ij}^{\, A} = V_\text{egg}(r_j, r_i, v_i)
        \label{eq:beta_a}
    \end{equation}
    Unlike the SFM’s repulsive potential \(V\) \cite{sfm95}, which assumes circular potentials with limited directional awareness, our modified potential, \(V_\text{egg}\) enhances directionality by stretching the potential along the agent's direction of motion into an egg shape. This formulation, \(V_\text{egg}(r_j, r_i, v_i)\), captures the strength of agent \(i\)'s potential at agent \(j\)'s position more effectively.

    Secondly, if the focal agent intrudes into agent \( j \)'s personal space, it actively triggers an interaction with \( j \). In this case, the focal agent monitors whether \( j \) reacts according to its expectation. Therefore, we assess how the strength of $j$'s potential at the position of $i$ changes over time as $i$ and $j$ move.
    This aspect of the interaction is formalized as $\beta_{ij}^{\,B}$ in \cref{eq:beta_b}, where $r_i^* = r_i + v_i \cdot  N_{DG} \, \Delta t$ represent the estimated future position of agent $i$, with $\Delta t$ denoting the time interval between state observations. 
    \begin{equation}
        \beta_{ij}^{\,B} = V_\text{egg}(r_i^*, r_j^*, v_j) - V_\text{egg}(r_i, r_j, v_j)
        \label{eq:beta_b}
    \end{equation}
    When the discretization step size \(N_{DG}\) is small, e.g., \(N_{DG} = 1\), \(\beta_{ij}^{\,B}\) can be interpreted as a discrete directed gradient of \(j\)'s repulsive potential along the direction of \(i\)'s velocity over time. 
    Hence, we name our prior DG-SFM. Finally, we combine the importance scores $\beta_{ij}^{\, A}$ and $\beta_{ij}^{\,B}$
    through a weighted sum and normalize the result using a softmax function over all neighbors \(J_i\) of \(i\), similar to SKGACN. This ensures that \(\beta_{ij} \in [0, 1]\) and \(\sum_{j \in J_i} \beta_{ij} = 1\).

\subsubsection{Interaction Prior Integration}
    After introducing an agent-to-agent attention layer and a class-agnostic prior, we address the challenge of effectively integrating this prior into attention mechanisms.
    We incorporate the prior during the agent-to-agent stage, as the attention scores in this layer are human-understandable \cite{zhang2022explainable} and located in the final encoder layer, where attention scores encode an interpretable measure of each agent’s relevance to the focal agent \cite{limeros2023towards}, thus maximizing their influence on the decoder’s predictions.
    
    To the best of our knowledge, no previous work has incorporated an interaction prior into an attention mechanism for mixed traffic trajectory prediction or adequately addressed its implications. Thus, inspired by integration methods outlined in \Cref{sec:sota:interaction_prior_integration}, we investigate two promising approaches to compare their impact on the attention scores' interpretability. 
    \paragraph{Multiply-and-Renormalize (MnR)} \textit{MnR} is a simple and easy-to-integrate approach. It combines the network's predicted attention scores \(\alpha_{ij}^\text{pred}\) 
    with the prior interaction importance scores \(\beta_{ij}\) 
    by multiplying them. The \textit{combined attention score} \(\alpha_{ij}^\text{cmb}\) for neighbor \(j\) of agent \(i\) is computed as per \cref{eq:mnr}. We view \(\alpha_{ij}^\text{cmb}\) as a trustworthy attention score guided by the prior. Consequently, the attention layer uses it in place of \(\alpha_{ij}^\text{pred}\) to fuse the embeddings of surrounding agents \(J_i\) with the focal agent's embedding.
    \begin{equation}
    \alpha_{ij}^\text{cmb} = \frac{\alpha_{ij}^\text{pred} \cdot \beta_{ij}} {\sum_k \alpha_{ik}^\text{pred} \cdot \beta_{ik}}
    \label{eq:mnr}
    \end{equation}
\paragraph{Gating-and-Loss (GnL)}
    \textit{GnL} is a more flexible approach for prior integration, inspired by the success of gating mechanisms in interpreting feature importance in NLP and computer vision \cite{margatina2019, ame}. It introduces a gating multi-layer perceptron (MLP) to compute gating values for agent \(i\), based on the agent embedding \(u_i\), the embeddings of its \(K\)-nearest neighbors \(u_j\), predicted attention scores \(\alpha_{i}^\text{pred} \in [0,1]^{|J_i|}\) 
    and the prior's interaction importance scores \(\beta_{i} \in [0,1]^{|J_i|}\), as shown in \cref{eq:sigm_gnl}. In this formulation, \(\text{sigm}\) is a sigmoid function, \(W \in \mathbb{R}^{|J_i| \times \text{hidden}}\) is the weight matrix, and \(b \in \mathbb{R}^{|J_i|}\) is the bias vector of the gating MLP. 
    \begin{equation}
        \sigma_{i} = \text{sigm} \left(
    W \cdot \text{concat}\left(u_i,\, \{u_j\}_{j \in J_i}, \, \alpha_{i}^\text{pred}, \,\beta_i\right)+ b
    \right)
        \label{eq:sigm_gnl}
    \end{equation}
    Subsequently, the combined attention score between agents \(i\) and \(j\) is computed as a weighted sum of the predicted attention and the prior, as shown in \cref{eq:gnl_sum}. The learned gating value \(\sigma_{ij}\) determines the extent to which the prior influences the attention layer's score. Finally, the combined attention scores are normalized across all neighbors \(j\).
    \begin{equation}
    \alpha_{ij}^\text{cmb} = \sigma_{ij} \cdot \alpha_{ij}^\text{pred} + (1 - \sigma_{ij}) \cdot \beta_{ij}
    \label{eq:gnl_sum}
    \end{equation}
    
    To complement the gating mechanism, we introduce an auxiliary loss that minimizes the Kullback–Leibler divergence between the combined attention scores \(\alpha_{i}^{\text{cmb}}\) and the prior scores \(\beta_{i}\), as illustrated in \cref{eq:gnl_loss}. The proposed auxiliary loss improves training stability and prevents the network from defaulting to a gating value \(\sigma_{ij}\) of 1, which would prioritize predicted attention scores and completely disregard the prior. For multi-head attention, we compute the average of this loss over all attention heads. This allows for variation in the heads while ensuring that they adhere to the prior when their results are put together.
    \begin{equation}
        \mathcal{L}_\text{attn}^{\,i} =  \frac{1}{|J_i|} \: D_{KL} \left( \beta_{i} \:\middle|\middle|\: \alpha_{i}^{\text{cmb}} \right)
        \label{eq:gnl_loss}
    \end{equation}

\subsection{Feasible Decoder}
\label{sec:method:dec}
    After encoding all agents and map elements with the proposed interpretable encoder, the network utilizes an anchor-based feasible decoder to predict multimodal trajectories for each focal agent \(i\) in the scene, followed by our prediction and confidence heads. To ensure kinematic feasibility, we employ action-space predictions with class-specific kinematic layers, following insights from \Cref{sec:sota:feasibility}. Separate prediction heads for each agent class enable the design of kinematic layers with distinct control inputs and feasibility constraints. Additionally, class-specific classification heads are introduced to give the network greater flexibility in distinguishing between agent classes and classifying their behaviors more independently.
\subsubsection{Kinematic Priors}
    We adopt the unicycle model, as defined in Trajectron++ \cite{trajectron++}, as the kinematic model for vehicles and cyclists. 
    Although the bicycle model has been more prevalent in trajectory prediction, the unicycle model is computationally more efficient, free of agent-specific parameters, and does not require online parameter estimation.
    Note that our approach could be expanded to any other set of kinematic models for any other taxonomy of agent classes.

\begin{figure}
        \centering
        \begin{subfigure}{0.33\linewidth}
          \centering
          \includegraphics[width=0.7\textwidth]{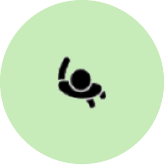}
          \vspace{0.05cm}
          \caption{Single Integrator}
        \end{subfigure}%
        ~
        \begin{subfigure}{0.33\linewidth}
          \centering
          \includegraphics[width=0.45\textwidth]{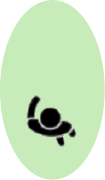}
          \caption{Double Integrator}
        \end{subfigure}%
        ~
        \begin{subfigure}{0.33\linewidth}
          \centering
          \includegraphics[width=0.75\textwidth]{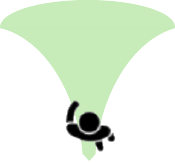}
          \vspace{0.05cm}
          \caption{Unicycle}
        \end{subfigure}
        \caption{
            The reachable area within a single time step when constraining a pedestrian walking with kinematic models. Among these, we find that the double integrator best represents real-world behavior.
        }
        \label{fig:kin_models}
    \end{figure}

    For pedestrians, there is no consensus in the literature on a standard kinematic model. Therefore, we compare two state-of-the-art models, the single integrator and the unicycle, both defined as in Trajectron++ \cite{trajectron++}. Additionally, we introduce a novel double integrator model, where the control inputs are \(x\)- and \(y\)-axis accelerations. An Euler integrator updates the agent's velocity and position at each time step based on these control inputs. As shown in \Cref{fig:kin_models}, the unicycle model overly restricts pedestrian motion, limiting curvature in an unrealistic manner. In contrast, the single integrator model is highly flexible but fails to account for acceleration limits. We find that the double integrator model provides the best trade-off between realism and constraints by allowing pedestrians to flexibly change heading while constraining velocity 
    based on acceleration limits.
\subsubsection{Kinematic Prior Integration}
\label{sec:method:dec:prior_integration}
Subsequently, the proposed kinematic models are incorporated into class-specific prediction heads. Instead of directly predicting trajectories in Cartesian space, each prediction head estimates control inputs recurrently, as described in DKM \cite{dkm}. These control inputs are then propagated through their respective kinematic model for each class to predict the future trajectories.
    
    To guarantee that the kinematic layer estimates only kinematically feasible trajectories, we impose feasibility limits on the control inputs predicted by the network.
    Acceleration and heading rate limits are enforced by applying a tanh function to the prediction head's outputs and rescaling them within the feasible bounds. For the unicycle and double integrator models, velocity limits are maintained by clipping the acceleration at each step to prevent exceeding the maximum velocity. For the single integrator, which does not account for accelerations, we directly apply the tanh function to the predicted velocities.
    
    We derive the control input limits based on physical constraints and empirical observations of vehicle and pedestrian behavior.
    For vehicles, we adopt the well-established limits of \([-8, 8]\) m/s\(^2\) for acceleration and \([-0.3, 0.3]\) 1/m for curvature \cite{dkm, ptnet}. For pedestrians, as prior studies lack specific limits, we draw on research into human walking and acceleration behavior \cite{zebala2012, graubner2009, fuerstenberg2005} to define appropriate constraints. These studies suggest that a reasonable maximum acceleration for pedestrians is \([-8, 8]\) m/s\(^2\) while their speed is limited to $[0, 10]$ m/s.

\section{Experiment Setup}
This section outlines experiments designed to evaluate whether our model improves the trustworthiness of trajectory prediction, with emphasis on feasibility and interpretability.
\subsection{Dataset}
    Our approach is evaluated on the Argoverse 2 (AV2) Motion Forecasting Dataset \cite{argoverse2}, a prominent benchmark for trajectory prediction with 250,000 urban traffic scenes. The dataset has a sampling interval of $\Delta t$ = 0.1 seconds, providing $T_o$ = 50 observation steps and $T_h$ = 60 prediction steps. The focal agent class distribution in AV2 is significantly imbalanced, comprising 92\% vehicles, 6\% pedestrians, and 2\% cyclists in both the training and validation splits \cite{argoverse2}.
\subsection{Implementation Details}
    Our models are trained with HPTR's position and confidence loss.
    We disabled HPTR's auxiliary speed, velocity, and yaw prediction, along with their respective losses, as these auxiliary outputs are predicted independently from the trajectories and may introduce contradictions.
    Instead, our network predicts these outputs consistently with the trajectory through our proposed kinematic models. Unlike HPTR, we do not use their negative log-likelihood loss function for position loss, as it would require propagating probability distributions through the kinematic layer—a direction of interest for future work but beyond this paper’s scope. Instead, we adopt the Huber loss from LaneGCN \cite{lanegcn}.   
    When integrating the attention prior via the \textit{GnL} method, we add the attention loss $\mathcal{L}_{\text{attn}}$ to training losses with a weight of 0.1. Our preliminary experiments showed that assigning equal weight to all losses caused the model to overemphasize reducing the attention loss $\mathcal{L}_{\text{attn}}$ to near-zero values while neglecting proper optimization of the position and confidence losses.
    To ensure comparability, we adopted HPTR's hyperparameter configuration. For our DG-SFM prior, we set the discretization step size \( N_{\text{DG}}\) to 10.
    Our final network, with all modifications, has 17.7M parameters versus HPTR's 15.2M. We trained all ablation networks for 15 epochs and trained the final network for 30 epochs. 

\subsection{Evaluation Metrics}
    We assess our approach across three metric categories: accuracy, interpretability, and feasibility.
    For accuracy, we rely on the popular benchmark metrics in \cite{womd, argoverse2}.
    For interpretability, we measure the difference between the prior's interaction importance score and the network's attention scores $\Delta \alpha(i) = \frac{1}{|J_i|} \sum_{j \in J_i} |\alpha_{ij} - \beta_{ij}| $.
    We then evaluate the correlation of minADE and $\Delta \alpha$.
    Depending on the prior integration method, we use either the predicted scores \(\alpha^\text{pred}_{ij}\) or the combined scores \(\alpha^\text{cmb}_{ij}\) as the attention scores, selecting the one with the stronger correlation.
    For feasibility, we evaluate the percentage of trajectory steps that violate feasibility limits specified in \Cref{sec:method:dec:prior_integration}. Finally, we calculate class-specific metrics by restricting the evaluation to predictions for a particular class of agents.

\section{Evaluation}
We evaluated our proposed modifications across all three metric categories. We begin with reasoning to determine the most effective interaction prior for our final model. Next, we assess the kinematic layers' impact on feasibility and select the pedestrian model that best reproduces the ground truth while balancing realism and feasibility. Finally, we conduct ablation studies based on these findings and compare our final model against other baselines.
\subsection{Reasoning}
\begin{table}
    \centering
    \caption{
            Correlation $\rho$ between the minADE and the prior-to-attention difference $\Delta \alpha$.
            As a baseline, we add an agent-to-agent transformer layer 
            to plain HPTR for capturing interactions, but without integrating any prior, referred to as HPTR$_i$.
            We compare HPTR$_i$ with several network configurations that integrate an interaction prior, referred to as HPTR$_{i+p}$.
    }
    \label{tab:eval:correlation}
\begin{tabular}{@{}lcc|c@{}}
\toprule
Model~~~~~~ & \begin{tabular}[c]{@{}c@{}}Prior Integration\\ Method\end{tabular} & Prior & $\rho\,$(minADE$_6$, $\Delta\alpha$) ($\uparrow$)\\ \midrule
\multirow{2}{*}{HPTR$_{i}$} & \multirow{2}{*}{-} & SKGACN & 0.03 \\
 &  & DG-SFM & 0.13 \\ \midrule
\multirow{4}{*}{HPTR$_{i+p}$} & MnR & SKGACN & 0.14 \\ \cmidrule(l){2-4} 
 & \multirow{3}{*}{GnL} & SKGACN & 0.22 \\
 &  & DG-SFM & \textbf{0.23} \\
 &  & L2 & -0.13 \\ \bottomrule
\end{tabular}
\end{table}
    
    In \Cref{tab:eval:correlation}, we investigate whether an integrated interaction prior improves the interpretability of the network's reasoning and compare various priors and integration methods.
    Particularly when comparing HPTR$_i$ attention scores to DG-SFM's prior scores, we observe a statistically significant correlation despite the network not being exposed to the prior during training. This correlation suggests a tendency for the network's prediction to be incorrect when the network's attention deviates significantly from the prior model's importance score. This tendency substantiates that aligning the encoder's reasoning with human reasoning can enhance the prediction's interpretability and accuracy.
    Thus, we conclude that minimizing the discrepancy between the model's attention and the prior is valuable. As larger differences coincide with incorrect predictions, guiding the network's attention using an interaction prior could improve the prediction.

    HPTR$_{i+p}$ models, which use a prior to guide the attention, improve the interpretability of the encoder's interaction reasoning by further strengthening the correlation. 
    This emerging correlation could serve as a foundation of a monitoring system, where significant deviations between the model's attention and the prior may indicate an unreasonable or incorrect prediction. Such an indication could, for example, prompt safety driver intervention in an SAE level 4 autonomous system.
    Comparing the integration methods MnR and GnL when using the same prior, GnL achieves a stronger correlation, offering better interpretability. Thus, we choose GnL for our final model.

    Comparing the prior models, both SKGACN and DG-SFM lead to a similarly strong correlation. On the other hand, integrating the simple L2 prior does not yield any meaningful correlation. The L2 prior assigns each agent an interaction importance score inversely proportional to its Euclidean distance from the focal agent.
    Since the L2 prior does not adequately represent the human intuition behind interaction importance, it does not facilitate the same correlation as the SKGACN and DG-SFM priors.
    DG-SFM resulted in a significantly stronger correlation when no prior is integrated. Furthermore, the mathematically substantiated advantages of its modeling, highlighted in \Cref{sec:method:enc:interaction_priors}, make it the more robust choice for complex scenarios.
    When DG-SFM guides our network's interaction attention, it fits human intuition on qualitative examples like \Cref{fig:dg-sfm_attn}, where HPTR failed.
    Hence, we choose DG-SFM over SKGACN for our final model.


\subsection{Prediction Feasibility}

\subsubsection{Infeasible Motions}
\label{sec:eval:feasibility:infeasibility}
\begin{table}
    \centering
    \caption{
        Infeasibility rate in the ground truth data of AV2's validation split, the trajectories predicted by HPTR \cite{hptr}, and our final model. 
        We deem a trajectory step infeasible if its acceleration or curvature exceeds the class-specific physical limits in \Cref{sec:method:dec:prior_integration}. 
        We deem a trajectory infeasible if any of its steps are infeasible.
    }
    \label{tab:eval:infeasibility}
\resizebox{\linewidth}{!}{%
\begin{tabular}{@{}l|ccc|ccc@{}}
\toprule
\multirow{2}{*}{Trajectories} & \multicolumn{3}{c|}{Infeasible Steps [\%]} & \multicolumn{3}{c}{Infeasible Trajectories [\%]} \\
 & Acceleration & Curvature & Any & Acceleration & Curvature & Any \\ \midrule
Ground Truth & 0.6\% & 8.7\% & 9.3\%  & 11.0\%  & 23.6\%  & 27.0\%  \\
HPTR Predictions & 3.4\% & 14.0\% & 17.0\% & 56.0\% & 42.0\% & 88.0\% \\
Our Predictions & \textbf{0\%} & \textbf{0\%} & \textbf{0\%} & \textbf{0\%} & \textbf{0\%} & \textbf{0\%} \\ \bottomrule
\end{tabular}
}
\end{table}
    \Cref{tab:eval:infeasibility} highlights the presence of significant physically infeasible steps in both the dataset and HPTR's predictions. Further discussion of the noise and infeasibility in AV2 can be found in~\cite{yao2023empirical}.
    This finding substantiates the rationale behind introducing our kinematic layer, which guarantees physical feasibility for every prediction. Without a kinematic layer, the network likely learns to reproduce these infeasible trajectories, effectively modeling sensor noise to minimize the distance to ground-truth trajectories. Enforcing kinematic constraints ensures that the network produces realistic, kinematically feasible predictions. This not only makes the predictions more trustworthy but also improves the model's generalizability. 

\begin{table}
	\centering
	\caption{
        Reproduction error introduced by kinematic models on AV2 validation split.
        Our greedy reproduction algorithm picks the best control inputs for infeasible steps to get as close as possible to the ground-truth trajectory's next position and heading.
	}
	\label{tab:eval:reproducability}
	\resizebox{\linewidth}{!}{%
\begin{tabular}{@{}ll|ccc@{}}
\toprule
Agent Class & Kinematic Model & minADE$_6$ & minFDE$_6$ & MR$_2$ \\ \midrule
All & (class-specific) & \textbf{0.206} & \textbf{0.574} & \textbf{2.2\%} \\ \midrule
\multirow{3}{*}{Pedestrian} & Single Integrator & 0 & 0 & 0 \\
 & Double Integrator & \textbf{$\boldsymbol{2 \cdot 10^{-4}}$} & \textbf{$\boldsymbol{2 \cdot 10^{-4}}$} & \textbf{0} \\
 & Unicycle & 0.521 & 1.14 & 6.2\% \\ \bottomrule
\end{tabular}
	}
\end{table}

\begin{table*}[!ht]
	\caption{
            Ablation study of the per-class accuracy of HPTR with different combinations of our proposed modifications. We compare these variants to the original HPTR model \cite{hptr} re-trained with the Huber loss for 15 epochs to ensure comparability. \(\vartriangleright\) denotes that the corresponding network uses a single agent-to-agent interaction layer shared across all agent classes.
	}
    \label{tab:eval:acc_ablation}
	\resizebox{\textwidth}{!}{%
\begin{tabular}{@{}lccc|cccc|cc|cc|cc@{}}
\toprule
\multirow{4}{*}{Model} & \multirow{4}{*}{\begin{tabular}[c]{@{}c@{}}Kinematic \\ Layer\end{tabular}} & \multirow{4}{*}{\begin{tabular}[c]{@{}c@{}}Class-Specific\\ Interaction\\ Layers\end{tabular}} & \multirow{4}{*}{\begin{tabular}[c]{@{}c@{}}Interaction\\ Prior\\ Integration\end{tabular}} & \multicolumn{4}{c|}{\multirow{2}{*}{All Agents}} & \multicolumn{2}{c|}{\multirow{2}{*}{Vehicle}} & \multicolumn{2}{c|}{\multirow{2}{*}{Pedestrian}} & \multicolumn{2}{c}{\multirow{2}{*}{Cyclist}} \\
 &  &  &  & \multicolumn{4}{c|}{} & \multicolumn{2}{c|}{} & \multicolumn{2}{c|}{} & \multicolumn{2}{c}{} \\
 &  &  &  & \multirow{2}{*}{\begin{tabular}[c]{@{}c@{}}Brier-\\ minFDE$_6$\end{tabular}} & \multirow{2}{*}{mAP ($\uparrow$)} & \multirow{2}{*}{minADE$_6$} & \multirow{2}{*}{minFDE$_6$} & \multirow{2}{*}{minADE$_6$} & \multirow{2}{*}{minFDE$_6$} & \multirow{2}{*}{minADE$_6$} & \multirow{2}{*}{minFDE$_6$} & \multirow{2}{*}{minADE$_6$} & \multirow{2}{*}{minFDE$_6$} \\
 &  &  &  &  &  &  &  &  &  &  &  &  &  \\ \midrule
Retrained HPTR &  &  &  & 2.28 & 0.275 & 0.82 & 1.49 & 0.96 & 1.73 & 0.44 & 0.80 & 1.07 & 1.93 \\ \midrule

\multirow{3}{*}{HPTR with} & \multicolumn{1}{l}{} & \multicolumn{1}{c}{\(\vartriangleright\)} & MnR \& SKGACN & 2.39 & 0.237 & 0.85 & 1.53 & 1.03 & 1.84 & 0.46 & 0.86 & 1.09 & 1.94 \\
 & \multicolumn{1}{l}{} & \multicolumn{1}{c}{\(\vartriangleright\)} & GnL \& SKGACN & \textbf{2.33} & 0.261 & 0.83 & 1.47 & \textbf{0.98} & \textbf{1.77} & 0.47 & 0.86 & 1.03 & 1.77 \\
 & \multicolumn{1}{l}{} & \multicolumn{1}{c}{\(\vartriangleright\)} & GnL \& DG-SFM & 2.36 & \textbf{0.273} & \textbf{0.82} & \textbf{1.46} & 1.00 & 1.79 & \textbf{0.47} & \textbf{0.84} & \textbf{1.00} & \textbf{1.73} \\ \midrule
 
\multirow{3}{*}{HPTR with} & \ding{51} &  &  & 2.58 & \textbf{0.231} & 1.05 & 1.80 & 1.21 & 2.04 & 0.49 & 0.88 & 1.46 & 2.48 \\
 & \ding{51} & \ding{51} &  & 2.50 & 0.179 & 0.99 & 1.70 & 1.18 & 1.95 & 0.49 & 0.89 & \textbf{1.31} & \textbf{2.27} \\
 & \ding{51} & \ding{51}& GnL \& DG-SFM & \textbf{2.40} & 0.230 & \textbf{0.97} & \textbf{1.67} & \textbf{1.03} & \textbf{1.85} & \textbf{0.47} & \textbf{0.85} & 1.42 & \textbf{2.27} \\ \bottomrule
\end{tabular}
}
\end{table*}
\begin{table*}
	\caption{
          Comparison of our final model, plain HPTR re-trained with the Huber loss, and the official HPTR model trained by Zhang et al. \cite{hptr}. We evaluate them on the AV2 validation split. The correlation $\rho\,$(minADE$_6$, $\Delta\alpha$) is not measurable for the two baselines, as they lack an agent-to-agent interaction layer.
	}
    \label{tab:eval:acc_sota}
	\resizebox{\textwidth}{!}{%
\begin{tabular}{l|cccccccccc}
\toprule
\multirow{2}{*}{Model} & \multirow{2}{*}{\begin{tabular}[c]{@{}c@{}}Brier-\\ minFDE$_6$\end{tabular}} & \multirow{2}{*}{mAP ($\uparrow$)} & \multirow{2}{*}{minFDE$_6$} & \multirow{2}{*}{minFDE$_1$} & \multirow{2}{*}{MR$_2$} & \multirow{2}{*}{minADE$_6$} & \multirow{2}{*}{minADE$_1$} & \multirow{2}{*}{\begin{tabular}[c]{@{}c@{}}Infeasible\\ Steps\end{tabular}} & \multirow{2}{*}{\begin{tabular}[c]{@{}c@{}}Infeasible\\ Predictions\end{tabular}} & \multirow{2}{*}{$\rho\,$(minADE$_6$, $\Delta\alpha$) ($\uparrow$)} \\
 &  &  &  &  &  &  &  &  &  &  \\ \midrule
Official HPTR  & \textbf{2.02} & \textbf{0.302} & \textbf{1.28} & \textbf{6.17} & \textbf{14.3\%} & \textbf{0.69} & \textbf{2.44} & 17.3\% & 87.7\% & - \\
Retrained HPTR & 2.28 & 0.275 & 1.49 & 6.57 & 17.4\% & 0.82 & 2.64 & 26.7\% & 99.6\% & - \\
Ours & 2.27 & 0.278 & 1.51 & 6.67 & 18.2\% & 0.90 & 2.69 & \textbf{0\%} & \textbf{0\%} & \textbf{0.12} \\ \bottomrule
\end{tabular}
}
\vspace{-0.4cm}
\end{table*}
\subsubsection{Accuracy of Kinematic Models} 
    The infeasible ground-truth trajectories in the dataset imply that a network enforcing physical feasibility cannot precisely reproduce some trajectories when adhering to physical limits.
    For example, the dataset contains trajectories with accelerations as high as 202 m/s$^2$ for vehicles and 31 m/s$^2$ for pedestrians, far exceeding physical limits. To address this, we assess how accurately our kinematic models can approximate the ground-truth trajectories, quantifying their reproduction error.
    In \Cref{tab:eval:reproducability}, we measure this error over all agent classes when using the unicycle model for vehicles and cyclists and the double integrator model for pedestrians, as defined in \Cref{sec:method:dec}.
    
    The presence of infeasible trajectories in the dataset establishes the reproduction error, discussed in \Cref{tab:eval:reproducability}, as the lower bound on prediction accuracy for networks adhering to kinematic limits. While networks without feasibility constraints may achieve higher metric scores by learning noisy or unrealistic behaviors, these predictions are less trustworthy and physically implausible. In contrast, our kinematic layer prevents the model from overfitting to noise, ensuring physically realistic and reliable predictions. We argue that a slight reduction in accuracy in exchange for improved feasibility and trustworthiness is a worthwhile trade-off, making the model more suitable for real-world deployment.
    Furthermore, we compare the reproduction errors introduced by the three kinematic models for pedestrians, as presented in \Cref{tab:eval:reproducability}. Among these, the double integrator model achieves the best balance between physical feasibility and versatility. It is more physically realistic than the single integrator, which lacks the ability to capture acceleration dynamics accurately. At the same time, unlike the unicycle model, the double integrator is flexible enough to reproduce nearly all of the dataset's trajectories without overly constraining the heading rate. This combination of feasibility and adaptability makes the double integrator model a robust choice, leading to its selection as a pedestrian model in our final network.
    
\subsection{Prediction Accuracy}
\subsubsection{Ablation Study}
\label{sec:eval:acc_ablation}
    First, we assess the impact of our modifications on the network's accuracy.
    \Cref{tab:eval:acc_ablation} further confirms our decision for GnL over MnR and DG-SFM over SKGACN, as the networks using these methods also lead to a higher accuracy. Moreover, the step-wise integration of our modifications shows that each layer enhances the model performance after an initial deterioration due to the kinematic layer, which reflects the accuracy limitations arising from the dataset's infeasible ground truth.
    
\subsubsection{Comparison to the State-of-the-Art}
    Finally, we compare our final model with the official HPTR model.
    As shown in \Cref{tab:eval:acc_sota}, the official HPTR model shows better accuracy than our model. However, this comparison is distorted by differences in the training loss function and epoch counts. The official HPTR uses a negative log-likelihood loss, which has an edge over our Huber loss in learning multimodal distributions through Gaussian position representation \cite{mtr}
    . However, adapting this loss for feasibility-constrained networks lies beyond the scope of this work.
    Moreover, the official HPTR's 75-epoch training period, compared to the 30 epochs of our model, gives it a notable advantage. Extending our training to 75 epochs was not viable due to the 30-day time requirement, which exceeded our resource constraints.
    
    Our ablation results for the different prior integration methods in \Cref{tab:eval:acc_ablation} closely matched plain HPTR, suggesting that the remaining performance gap is almost exclusively due to the kinematic layer. However, this trade-off is well justified. As discussed in \Cref{sec:eval:feasibility:infeasibility}, the kinematic layer prevents the model from learning unrealistic and noisy patterns present in the dataset.
    On balance, our model outweighs its minor accuracy disadvantage with two crucial benefits: 
    more trustworthy reasoning through interpretable interaction attention,
    and more trustworthy prediction through guaranteed physical feasibility.

\section{Conclusion}
Deep learning has advanced trajectory prediction, but its data-driven nature often results in untrustworthy predictions. Hybrid methods improve feasibility and interpretability by integrating kinematic and interaction priors. However, the literature lacks an interaction prior applicable to mixed traffic and lacks consensus on a pedestrian kinematic model that best represents real-world behavior. This paper advances the trustworthiness of trajectory predictions with two key layers: an agent-to-agent interaction embedding layer guided by the novel rule-based DG-SFM prior for interpretability and kinematic layers that enforce feasibility by converting predicted control inputs into kinematically viable trajectories. Experiments reveal that HPTR relies on interactions misaligned with human reasoning, which correlate with erroneous predictions. Integrating the DG-SFM prior improves alignment with human intuition and strengthens the correlation between prediction errors and deviations from prior-based importance scores. DG-SFM outperforms SKGACN and L2 priors. Additionally, analysis shows significant infeasible noise in the dataset and HPTR predictions, with tailored kinematic models enhancing feasibility across agent classes. The double integrator model, in particular, achieves superior pedestrian trajectory reproduction.

{
    \bibliographystyle{IEEEtran}
    \bibliography{references}
}

\end{document}